\documentclass{article}
\usepackage{spconf,amsmath,graphicx}

\def\x{{\mathbf x}}
\def\L{{\cal L}}

\usepackage[utf8]{inputenc} 
\usepackage[T1]{fontenc}    
\usepackage{hyperref}       
\usepackage{url}            
\usepackage{booktabs}       
\usepackage{amsfonts}       
\usepackage{nicefrac}       
\usepackage{microtype}      

\usepackage{subfigure} 
\usepackage{graphicx} 
\usepackage{mathtools} 
\usepackage{amssymb}
\usepackage{multirow}
\usepackage{algorithm}
\usepackage[noend]{algpseudocode}

\usepackage{xcolor}
\usepackage{footnote}
\usepackage{times}
\usepackage{epsfig}
\usepackage{amsmath}
\usepackage{fixltx2e}
\usepackage{color}
\usepackage{algpseudocode}

\usepackage[font=small,labelfont=bf]{caption}

\graphicspath{{./figures/}{./figures/draw/}{./figures/fobj_eval/}}

\def\X{{\mathbf X}}
\def\Y{{\mathbf Y}}
\def\V{{\mathbf V}}
\def\P{{\mathbf P}}
\def\Q{{\mathbf Q}}
\def\L{{\mathbf L}}
\def\A{{\mathbf A}}
\def\B{{\mathbf B}}
\def\I{{\mathbf I}}
\def\M{{\mathbf M}}

\def\D{{\mathbf D}}
\def\W{{\mathbf W}}
\def\0{{\mathbf 0}}

\def\x{{\mathbf x}}

\def\u{{\mathbf u}}

\def\z{{\mathbf z}}

\def\R{{\mathbb R}}

\ninept

\title{Simultaneous Low-Rank Component and Graph Estimation for High-Dimensional Graph Signals: Application to Brain Imaging}
\name{Liu Rui,  Hossein Nejati, Seyed Hamid Safavi, Ngai-Man Cheung}
\address{Singapore University of Technology and Design (SUTD), Singapore 487372}
\begin{document}
%
\maketitle
\begin{abstract}

We propose an algorithm to uncover the intrinsic low-rank component of a high-dimensional, graph-smooth and grossly-corrupted dataset, under the situations that the underlying graph is unknown.
Based on a model with a low-rank component plus a sparse perturbation, and an initial graph estimation, our proposed algorithm simultaneously learns the low-rank component and refines the graph.
The refined graph improves the effectiveness of the graph smoothness constraint and increases the accuracy of the low-rank estimation.
We derive the learning steps using ADMM.
Our evaluations using synthetic and real brain imaging data in a supervised classification task demonstrate encouraging performance.

\end{abstract}
\section{Introduction}
\label{sec:intro}

We consider the problem of uncovering the intrinsic low rank component of a high-dimensional dataset.  We further focus on data that resides on a certain graph and the data changes smoothly between the connected vertices \cite{shuman2013emerging}.  In many problems, the underlying graph is unknown or inexact \cite{frossard2015,ortega:2016,Kalofolias:2016}.  For example, the graph may be estimated from the input data which is grossly corrupted.  We propose an algorithm to estimate the low-rank component of the data, using the graph smoothness assumption to assist the estimation.  Our algorithm also simultaneously and iteratively refines the graph to improve the effectiveness of the graph smoothness constraint, thereby increasing the quality of low-rank component estimation.

High-dimensional data is common in many engineering areas such as image / video processing, biomedical imaging, computer networks and transportation networks.  
We are specifically interested in automatic analysis of brain imaging data. In many cases, the task is to find the spatiotemporal neural signature of a task, by performing classification on cortical activations evoked by different stimuli \cite{hossein2016, Nejati2015}. 
Common brain imaging techniques are Electroencephalography (EEG) and 
Magnetoencephalography (MEG).  
These measurements are high-dimensional spatiotemporal data.  For instance, in our experiments, we use 
a recumbent Elekta MEG scanner with 
306 sensors to record the brain activity for 1100 milliseconds.
Furthermore, the measurements are degraded by various types of noise (e.g., sensor noise, ambient magnetic field noise, etc.) and the noise model is complicated  (potentially non-Gaussian).
The  high-dimensionality and noise  limit both the speed and accuracy of the signal analysis, that may result in unreliable signature modeling for classification.  The high-dimensionality of these signals also increases the complexity of the classifier. 

Note that
it has been recognized that there are patterns of anatomical links, or  statistical dependencies or causal interactions between distinct units within a nervous system~\cite{bullmore2009complex}.  
Some techniques have also been developed to estimate this {\em brain connectivity graph}~\cite{hyde2012cross,brovelli2004beta}.  However, this task is complicated and in many cases the estimated graph may not be accurate.  Our contribution is to develop a robust algorithm to determine the reduced 
dimensionality components 
that include task-related information, under the assumption that the brain imaging data is graph-smooth but the knowledge of the graph is imperfect.
Specifically, 
our contributions are: (i) based on a model with a low-rank component plus a sparse perturbation, and an initial graph estimation,  we propose an algorithm to simultaneously learn the low-rank component and the graph; (ii) we derive the learning steps using ADMM~\cite{boyd2011distributed}; (iii) we evaluate the algorithm using synthetic and real brain imaging data in a supervised classification task.


\subsection{Related Work}

This work is inspired by \cite{frossard2015}.  While the focus of \cite{frossard2015} is to learn the connectivity graph topology, their algorithm also estimates some noise-free version of the input data as by-product.  Gaussian noise model and Frobenius norm optimization are employed in \cite{frossard2015}.  Therefore, their work is suitable for problem when noise is small.  In our work, starting from a model with a low-rank component plus a sparse perturbation, and an initial graph estimation, we adopt the idea of \cite{frossard2015} to incrementally refine the underlying connectivity graph, thereby better low-rank estimation of the data can be obtained.  As will be shown in our experiment, our method can perform better with high-dimensional graph data {\em grossly} corrupted by complicated noise, such as brain imaging signals.
In addition to \cite{frossard2015}, learning a graph from smooth signals has attracted a fair amount of interests recently \cite{ortega:2016,Kalofolias:2016}.  These works focus on learning the graph, and advanced formulations (e.g., matrix optimization problem) have been derived.  Estimation of the brain connectivity graph using a Gaussian noise model has been proposed in \cite{lu:2013}.  On the other hand, focusing on low-rank estimation, some works have proposed to incorporate spectral graph regularization~\cite{Jian13b,Shah15a,Shah16b}.  Their graphs are {\em fixed} in their algorithms, pre-computed from the noisy input data.  On the contrary, our algorithm uses the improved low-rank estimation to refine the graph, which in turn is used to improve the quality of the low-rank estimation.  Besides, 
graph signal processing has 
been applied to a few different brain imaging tasks.
In \cite{huang2016graph}, graph Fourier transform is applied to decompose brain signals into low, medium, and high frequency components for analysis of functional brain networks properties.
\cite{rui2016dimensionality} uses the eigenvectors of the graph Laplacian to approximate the intrinsic subspace of high-dimensional brain imaging data.  They experimented different brain connectivity estimations to compute the graph.  
\cite{behjat2015anatomically} presents a graph based framework for fMRI brain activation mapping.
Graph signal processing has
also been shown to be useful in  image compression \cite{hu2015multiresolution}, temperature data \cite{frossard2015}, wireless sensor data \cite{egilmez2014spectral}.  A few signal features motivated by graph signal processing have also been proposed~\cite{dong2013inference, kang2014complex}. 
Moreover, several linear / nonlinear dimensionality methods have been proposed that make use of the graph Laplacian of the sample affinity graphs~\cite{Belk03,He:2003,yan:2007}  These methods are geometrically motivated, aim to preserve the local structures of the data, and involve different algorithms compared to our work.



\section{SIMULTANEOUS LOW RANK AND GRAPH ESTIMATION}
\label{sec:lowrank_GraphEstimate}

We consider   $\X=(\x_1,\dots,\x_n) \in \R^{p \times n}$,
the high dimensional data matrix
 that consists of $n$ $p$-dimensional data points. 
For our brain imaging data, 
$\X$ are the measurements by the $p$ sensors at the $n$ time instants, i.e., $p$ time series.
We assume the data points have low intrinsic dimensionality and lie near some low-dimensional subspace.  We assume the following mathematical model for the data: 
\begin{equation}
\X = \L_0 + \M_0
\end{equation}
$\L_0 \in \R^{p \times n}$ is the low-rank component of the data matrix that is of primary interest in this paper, and $\M_0  \in \R^{p \times n}$ is a perturbation matrix.  We assume that $\M_0$ can have arbitrarily large magnitude but its support is sparse.  

Principal component analysis (PCA) is the most popular technique for determining the low-rank component with application domains ranging from image, video, signal, web content, to network. 
The \emph{classical PCA} finds the projection $\Q^T \in \R^{k \times n}$ of $\X$ in a $k$-dimensional ($k \le p$) linear space characterized by an orthogonal basis $\V \in \R^{p \times k}$, by solving the following optimization:
\begin{equation}
\begin{aligned}
& \underset{\V,\Q}{\text{minimize}}
& & \| \X-\V\Q^T\|_F^2 \\
& \text{subject to}
& & \V^T \V = \I
\end{aligned}
\end{equation}
The $\V$ and $\Q^T$ matrices are known as principal components and projected data points, respectively. $\L = \V\Q^T \in \R^{p \times n}$ is the approximation of the
low-rank component.
The classical PCA suffers from a few disadvantages.  
First, 
it is susceptible to grossly corrupted data in $\X$.
Second, it does not consider the implicit data manifold information.

Candes \emph{et al.} \cite{Cand11} addressed the first issue by designing \emph{Robust PCA}, which is robust to outliers by directly recovering the low-rank matrix $\L$ from the grossly corrupted $\X$:
\begin{equation}
\begin{aligned}
& \underset{\L,\M}{\text{minimize}}
& & \|\L\|_* + \delta \|\M\|_1 \\
& \text{subject to}
& & \X = \L + \M
\end{aligned}
\label{eq:rpca}
\end{equation}
$\|.\|_*$ denotes the nuclear norm which 
is used as a convex surrogate of rank.

In this work we propose to extend (\ref{eq:rpca}) with an additional graph smoothness regularization, while the underlying graph topology that captures the data correlation could be  {\em unknown} or {\em inexact} (thus some refinement is needed):
\begin{equation}
\begin{aligned}
& \underset{\L,\M, \Phi_f}{\text{minimize}}
& & \|\L\|_* + \delta \|\M\|_1 +   \gamma tr(\L^T \Phi_f \L) + \beta \|\Phi_f \|_F^2\\
& \text{subject to}
& & \X = \L + \M, \\
&&& \Phi_f \in \mathcal{L}
\end{aligned}
\label{eq:rpca-ug}
\end{equation}
Here $\Phi_f$ is the graph Laplacian of the feature graph  $\mathcal{G}$ describing the correlation between individual features: $\mathcal{G} = (\mathcal{V},\mathcal{E}, \W)$ consists of a finite set of vertices $\mathcal{V}$, with $|\mathcal{V}| = p$, a set of edges  $\mathcal{E}$, and a weighted adjacency matrix $\W = \{ W_{i,j} | W_{i,j} \ge 0 \}$, with $W_{i,j}$ quantifying the similarity between the $i$-th and $j$-th features of the $p$-dimensional measurement vectors. $\Phi_f = \D-\W$, with 
$\D$ being the diagonal degree matrix.
$\mathcal{L}$ is the set of all valid $p \times p$ graph Laplacian $\Phi$:
\begin{equation}
\mathcal{L}= \{ \Phi : \Phi_{ij} = \Phi_{ji} \leq 0, \Phi_{ii} = - \sum_{j \neq i} \Phi_{ij}  \}
\end{equation}

As will be further discussed in Section~\ref{sec:illust},
we solve (\ref{eq:rpca-ug}) iteratively using alternating minimization with the following justifications:
\begin{itemize}
\item
{\bf $\L, \M$ given $\Phi_f$:} 
For a given $\Phi_f$ (even a rough estimate),   $tr(\L^T \Phi_f \L)$  imposes an additional constraint on the underlying (unknown) low-rank data $\L$.  Specifically, 
\begin{equation}
tr(\L^T \Phi_f \L) = \frac{1}{2} \sum_{i,j}  W_{i,j} \|    \tilde{l_i} -  \tilde{l_j}    \|^2,
\end{equation}
where $\tilde{l_i} \in \R^n$ is the $i$-th row vector of $\L$.
Therefore, $tr(\L^T \Phi_f \L)$ in (\ref{eq:rpca-ug}) forces the row $i$ and $j$ of $\L$ to have similar values if $W_{i,j}$ is large.  Note that in our brain imaging data, individual rows represent the time series captured by sensors. Thus, $tr(\L^T \Phi_f \L)$ forces the low-rank  representations of the time series to be similar for highly correlated sensors.  Prior information regarding measurement correlation (such as the physical distance between the capturing sensors) can be incorporated as the initial $\Phi_f$  to bootstrap the estimation of $\L$. 
\item
{\bf $\Phi_f$ given $\L, \M$:}
On the other hand, for a given estimate of the low-rank data $\L$, $tr(\L^T \Phi_f \L)$ guides the refinement of $\Phi_f$ and hence the underlying connectivity graph $\mathcal{G}$. In particular, a graph $\mathcal{G}$ that is consistent with the signal variation
in $\L$ is favored: large $W_{i,j}$ if row $i$ and $j$ of $\L$ have similar values.  In many problems, the given graph for a problem can be noisy (e.g., the graph is estimated from the noisy data itself \cite{Shah15a,Shah16b}).  The proposed formulation iteratively improves $\Phi_f$ using the refined low-rank data.  The improved $\Phi_f$
 in turn facilitates the low-rank data estimation. 
\end{itemize}


\section{Learning Algorithm}
\label{sec:illust}

We propose to solve the problem in Eq (\ref{eq:rpca-ug}) with alternating minimization scheme where, at each step, we fix one or two variables and update the other variable.  

At the first step, for a given $\Phi_f$, we solve the following optimization problem using ADMM\cite{boyd2011distributed} with respect to $\L$ and $\M$. It means given a graph, it estimates the low rank matrix:
\begin{equation}
\begin{aligned}
& \underset{\L,\M}{\text{minimize}}
& & \|\L\|_* + \delta \|\M\|_1 +   \gamma tr(\L^T \Phi_f \L)\\
& \text{subject to}
& & \X = \L + \M, \\
\end{aligned}
\label{eq:rpca-ug-step1}
\end{equation}

At the second step, $\L$ and $\M$ are fixed and we solve the following optimization problem with respect to $\Phi_f$, which means that based on the low rank matrix we got in the previous step, it updates the graph.
\begin{equation}
\begin{aligned}
& \underset{\Phi_f}{\text{minimize}}
& & \gamma tr(\L^T \Phi_f \L) + \beta \|\Phi_f \|_F^2 \\
& \text{subject to}
& & \Phi_f \in \mathcal{L} \\
\end{aligned}
\label{eq:rpca-ug-step2}
\end{equation}

For equation (\ref{eq:rpca-ug-step2}), it can be written as:
\begin{equation}
\begin{aligned}
& \underset{\Phi_f}{\text{minimize}}
& & \gamma tr(\L^T\Phi_f\L)+\beta\|\z\|_F^2\\
& \text{subject to}
& & \Phi_f -\z = 0, \\
&&& \Phi_f \in \mathcal{L}
\end{aligned}
\label{eq:rpca-ug-step2-ADMM}
\end{equation}
We can form the augmented Lagrangian of (\ref{eq:rpca-ug-step2-ADMM}) as:
\begin{equation}
\begin{aligned}
L_\rho(\Phi_f,\z,\u) = &\gamma tr(\L^T\Phi_f\L)+\beta\|\Phi_f\|_F^2 \\
& +\frac{\rho}{2}\|\z-\Phi_f\|_F^2 + \langle \u,\z-\Phi_f \rangle \\
\end{aligned}
\label{eq:rpca-ug-step2-Lagrangian}
\end{equation}
Then we can get the following formula to update for $\Phi_f$, $\z$ and $\u$:

\begin{equation}
\begin{aligned}
& \Phi_f^{k+1} := \frac{\rho \z^k - \gamma \L^T\L + \u}{\frac{\beta}{2} +\rho}\\
& \z^{k+1}:= \prod_{\mathcal{L}}(\Phi_f^{k+1} - \frac{1}{\rho}\u^k)\\
& \u^{k+1}:= \u^{k} + \frac{1}{k}(\z^{k+1}-\Phi_f^{k+1}) \\
\end{aligned}
\label{eq:rpca-ug-step2-update}
\end{equation}

where $\rho>0$ is the Lagrangian parameter and $\prod_{\mathcal{L}}$ is the Euclidean projection onto set $\mathcal{L}$.

\section{Experiment}
\label{sec:Experiment}
\subsection{Synthetic Experiment}
\label{ssec:Synthetic}
In this synthetic experiment, we generate low-rank, graph-smooth and grossly-corrupted data.
We generate synthetic data with the following model:
\begin{equation}
\begin{aligned}
\X = \L_0 + \M_0
\end{aligned}
\label{eq:data-model}
\end{equation}
where $\L_0\in \R^{p\times n}$ is a low rank matrix with rank $r$ and $\M_0$ is the sparse matrix. 

We generate $\L_0$ as a product $\L_0 = \P\Y^T$ where $\P\in \R^{p\times r}$ and $\Y\in \R^{n\times r}$. $L_0$ is also graph-smooth and generated as follows.   The (ground-truth) graph consists of $p$ nodes, with each pair of nodes having a probability of $q$ to be connected together. The edge weights between different nodes are drew uniformly from 0 to 1 and presented in a $p\times p$ symmetric adjacency matrix $\W$. We calculate the Laplacian matrix $\Phi_f$ from $\W$ and compute the eigenvectors and eigenvalues of $\Phi_f$. The eigenvectors, corresponding to top $r$ eigenvalues, are selected as the columns of $\P$. For matrix $\Y$, the entries are independently sampled from a $N(0,1/p)$ distribution. Therefore, $\L_0$ is low-rank and graph-smooth. We introduce $k= \|\M_0\|_0 / p^2$ errors in the matrix $\M_0$ from an i.i.d Bernoulli distribution. Each corrupted entry takes a value $\pm1$ with a probability $k/2$. 


We compare the proposed method, GL-SigRep\cite{dong2014learning}, RPCAG\cite{Shah15a}, RPCA\cite{Cand11}, and PCA on the data to estimate the low rank matrix and the graph matrix. The estimation accuracies are evaluated by the reconstruction errors: $\| \hat{\L} -\L_0\|_F/\|\L_0\|_F$ and $\| \hat{\Phi_f} -\Phi_f\|_F/\|\Phi_f\|_F$. 
All the methods are initialized with the {\em same} feature similarity graph (consider each row of $\X$ as a node) computed using the procedure in \cite{Shah16b} with a K-nearest  neighbor strategy ($K = 10$). 


Table \ref{tab:SynResult} shows the  results on synthetic data generated by the Eq (\ref{eq:data-model}) with $p=30, n =50, r = 3, k =40\%, q =0.2$. With cross-validation, we set $\delta = \frac{2.5}{\sqrt{50}},\gamma = 1.5,\beta = 1.5$. The low rank approximation of proposed method achieves the smallest error. For the estimate graph matrix, the proposed method also achieves the smallest estimation error. Synthetic experiment results show that the proposed method can achieve good performance on extracting low rank approximation and the underlying graph from non-Gaussian noisy data.

\begin{table}\small 
	\begin{tabular}{|*{1}{p{70pt}|}*{2}{c|}}
		\hline
		Methods&Low Rank Matrix&Graph Matrix\\
		\hline
		Proposed Method& \textbf{0.4278} & \textbf{0.3325} \\
		\hline
		GL-SigRep&7.3263&1.4408\\
		\hline
		RPCAG&0.4657&-\\
		\hline
		RPCA&3.5883&-\\		
		\hline
		PCA&7.2942&-\\		
		\hline
	\end{tabular}
	\centering
	\vspace{+0.1cm}
	\normalsize
	\caption{Comparison of low rank matrix error and estimated graph error for synthetic data. }
	\label{tab:SynResult}
\end{table}

\subsection{Brain Imaging Data Experiment}
\label{ssec:BrainData}
We also apply our proposed method on a high-dimensional brain imaging dataset to extract the brain connectivity graph and the low rank approximation from the high dimensionality. This is practically useful for brain imaging studies: due to the high dimensionality of data, low signal-to-noise ratio, and small number of available samples, it is challenging to estimate the low rank approximation in these studies. 

The brain imaging dataset used here is a set of magnetoencephalography (MEG) signal recordings of brain activities, in response to two categories of visual stimuli: 320 face images and 192 non-face images. These images were randomly selected and displayed passively with no task, and 16 subjects were asked to simply fixate at the center of the screen. All images were normalized for size and brightness among other factors, and were each displayed once for 300 ms with random inter-stimulus delay intervals of 1000 ms to 1500 ms. We used a recumbent Elekta MEG scanner with 306 sensors to record the brain activity for 1100 milliseconds (100 milliseconds before and 1000 milliseconds after the presentation) for each stimuli. The classification task in this experiment is to distinguish signals evoked by face images from signals evoked by non-face images.

\subsubsection{Initial graph matrix}
\label{sssec:InitializationGraph}
In the proposed method, a suitable starting point is important for solving the optimization problem. We therefore initialize $\Phi_f$ with the brain connectivity matrix generated with the resting state measurements. The resting state in our experiment is 100ms of signal recording \emph{before} the stimuli presentation. 
Note that our method and all other methods are initialized with the same connectivity matrix.

Three different types of brain connectivity graphs are commonly used in the literature: structural connectivity, functional connectivity and effective connectivity. Structural connectivity shows the anatomical structure in the brain; functional connectivity quantifies functional dependencies between different brain regions; and effective connectivity shows directed or causal relationship between different brain regions~\cite{bullmore2009complex}. 

In this paper we use a coherence connectivity, a functional connectivity, quantifying oscillatory interdependency between different brain regions~\cite{rui2016dimensionality}. It is the frequency domain analog of the cross-correlation coefficient. Given two series of signals $a_t$, $b_t$ and a frequency $f$, the first step is to spectrally decompose the signal at target $f$ to obtain the instantaneous phase at each time point~\cite{schmidt2014whole}. After band-pass filtering each signal between $f\pm5Hz$, the convolution of $f(t)$ with a Morlet wavelet centered at frequency $f$ provides the instantaneous phase at time $t$. Thus, the two signals can be represented as: $a=E_a(t)e^{j\psi_a(t)}$ and $b=E_b(t)e^{j\psi_b(t)}$, where $E_a(t)$ and $E_b(t)$ are amplitudes. $\psi_a(t)$ and $\psi_b(t)$ are the phase for $\A$ and $\B$ at time t. Then the coherence connectivity edge is calculated as below:

\vspace*{-0.5\baselineskip}
\begin{equation}\label{coh}
w_{A,B}=\left|\frac{\frac{1}{T}\sum_{t=1}^{T}E_a(t)E_b(t)e^{j[\psi_a(t)-\psi_b(t)]}}{\sqrt{\frac{1}{T}\sum_{t=1}^{T}E_a(t)^2}\cdot\sqrt{\frac{1}{T}\sum_{t=1}^{T}E_b(t)^2}}\right|
\end{equation}

After we get the adjacency matrix $\W$, we can calculate the Laplacian matrix $\Phi_f = \D - \W$ as the initialization.



\subsubsection{Brain Imaging Data Experiment Results}
\label{sssec:BrainResult}
To evaluate the performance of proposed method, we use a supervised classifier, SVM, to classify the low-rank outputs into face and non-face classes. We compare these methods based on their classification accuracies as well as compatibility of their connectivity graph matrix to the related neuroscience findings on suggested cortical regions involving face processing.

MEG and EEG components corresponding to the face/non-face distinction have been reported at latencies of approximately 100 ms, and more reliably at 170 ms (also known as N170 marker, reported at about 145ms in MEG studies), after visual stimulus onset (e.g. see \cite{Perrett1987, Thorpe1996, Liu2002, Desimone2006}). In this experiment, we therefore choose the data from two time slots, namely 96ms to 105ms and 141ms to 150ms after the stimuli presentation, to be able to compare our automate assessment with the related neuroscience literature.

We applied commonly employed mean subtraction on the time-locked data. In addition to this step, we refrained from providing any prior information to our method. This is due to the goal of this paper to showcase the capabilities of our proposed method, and therefore chose a pure data driven approach to the problem of estimation of underlying connectivity graph. For example, we did not enforce constraints in the optimization step on the sensor sensitivity to field spread (i.e. nearby electrodes record similar brain activities). Similarly, we did not differentiate between the magnetometers and gradiometers. With no prior information fed to the method along-side the data, the estimated underlying graph can be easily compared with the results of other methods. This is while more specific experiments can be conducted in the future to reveal the effects of prior information on the results.

Given the input data and the initial graph, our proposed method outputs some low rank estimation $\L$.
We decompose the low rank matrix $\L$ using SVD,  select the components according to the rank of $\L$, project the data onto the components to obtain low-dimensionality representations,  and classify the reduced dimension data. We compare the proposed method with GL-SigRep, RPCAG, RPCA and PCA. Using cross-validation with $\delta \in \frac{[1:0.5:8]}{\sqrt{50}}$, $\gamma \in 10^{[-2:1:1]}$, $\frac{\beta}{\gamma} \in [1:5:10]$ (for step 2, only the ratio of $\beta$ and $\gamma$ matters the results), we set $\delta = \frac{1}{\sqrt{50}}, \gamma = 0.1, \beta = 0.5$ for the first time slot and $\delta = \frac{1}{\sqrt{50}}, \gamma = 0.01, \beta = 0.05$ for the second time slot.Table~\ref{tab:BrainResult} shows the classification results for different methods on the two time slots. The proposed method gives the best results for both time slots.

\begin{figure}
\begin{centering}
\begin{tabular}{cccc}
\multicolumn{4}{c}{\includegraphics[width=0.21\columnwidth]{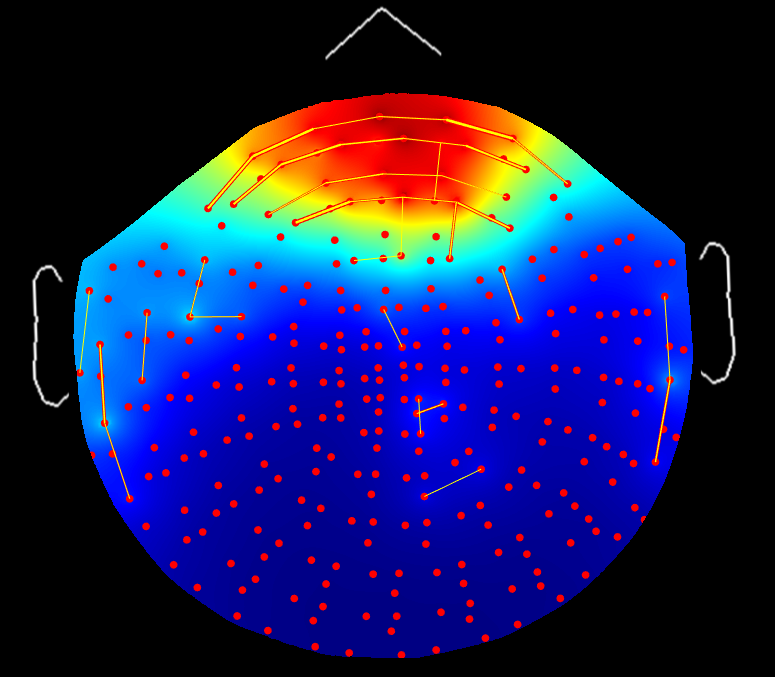}}\tabularnewline
\includegraphics[width=0.21\columnwidth]{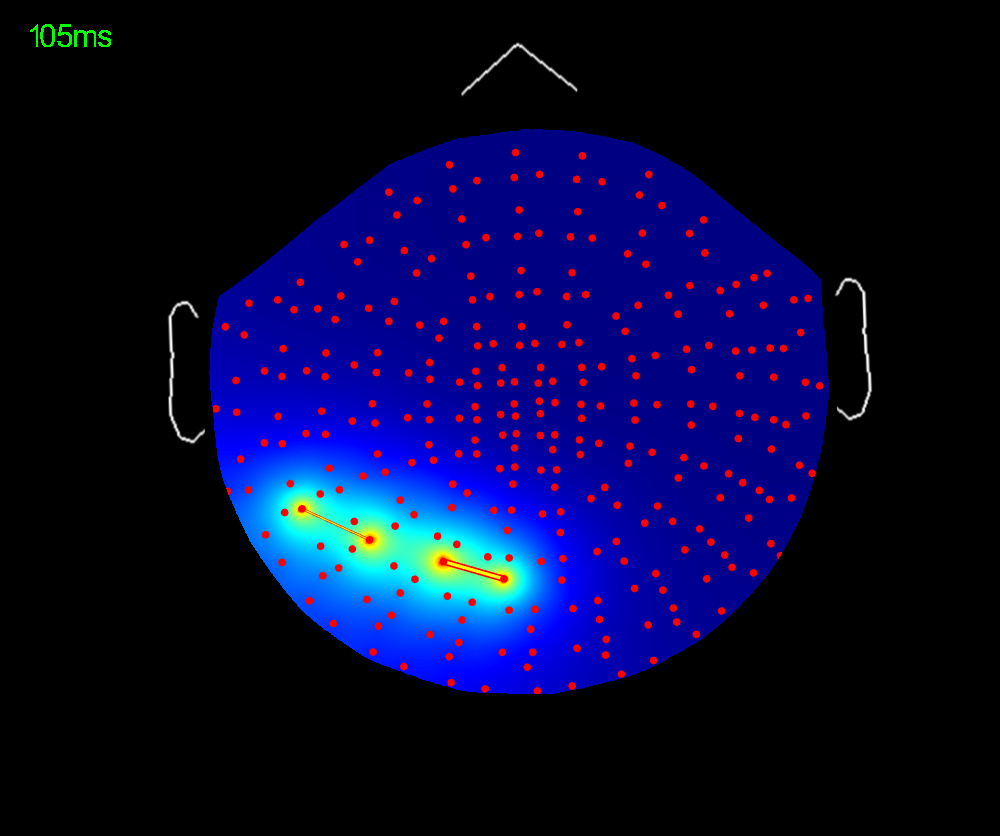} & 
\includegraphics[width=0.21\columnwidth]{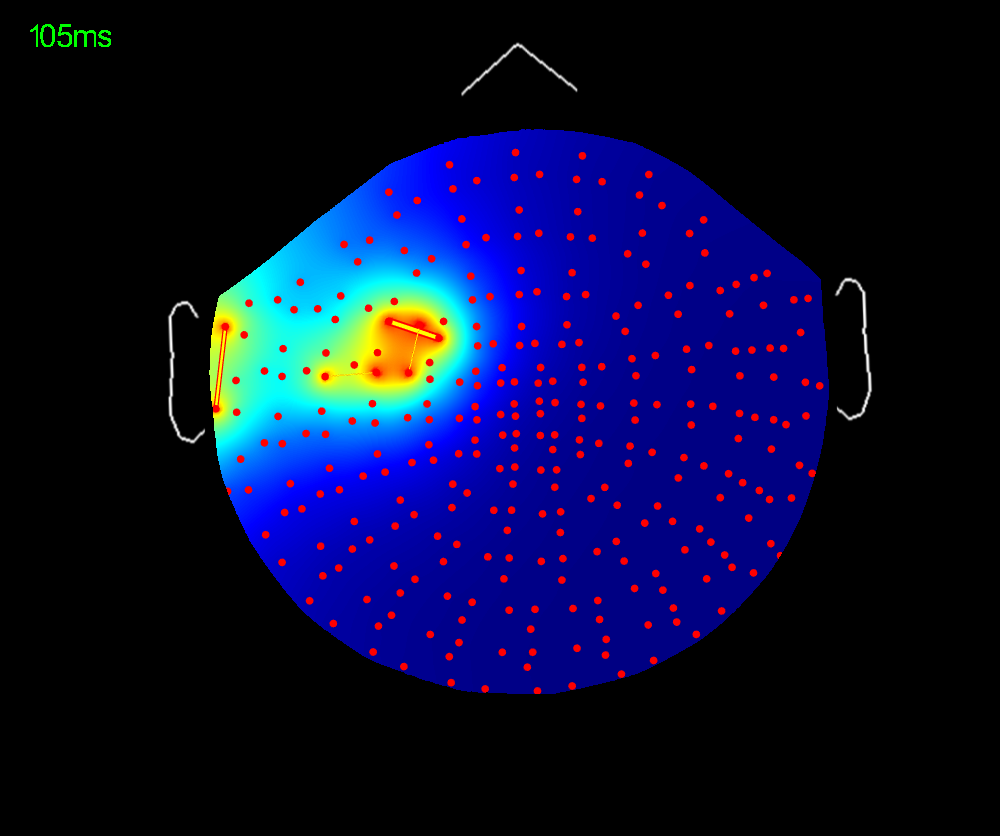}&
\includegraphics[width=0.21\columnwidth]{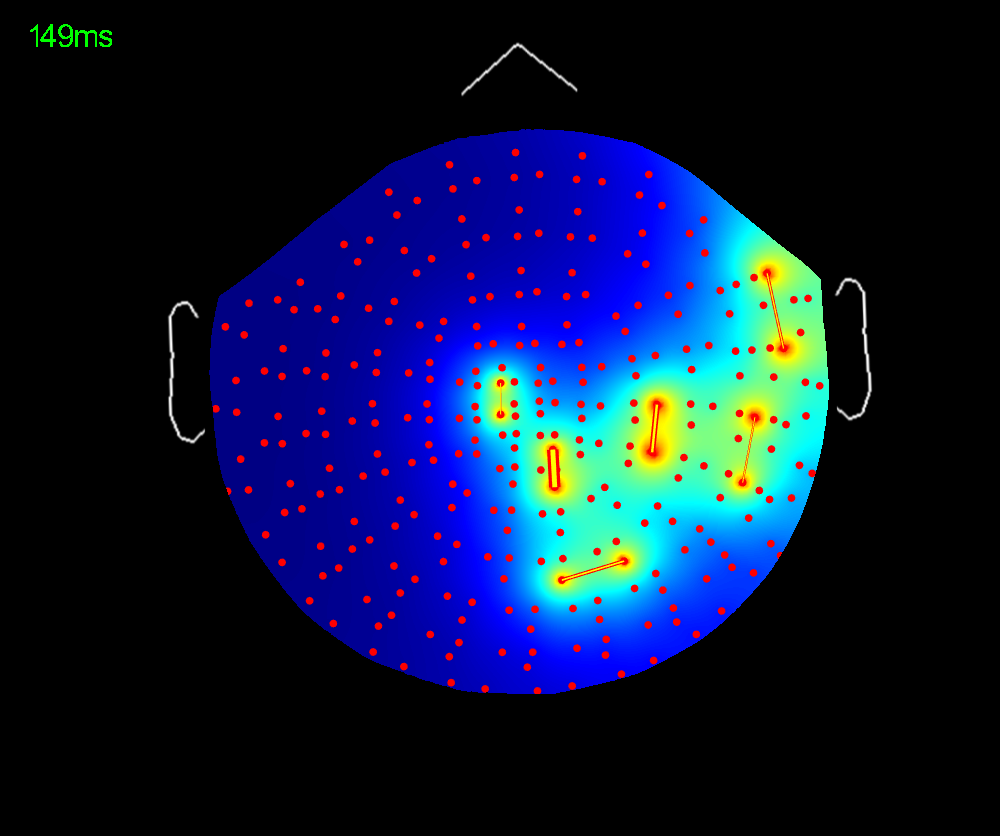} &
\includegraphics[width=0.21\columnwidth]{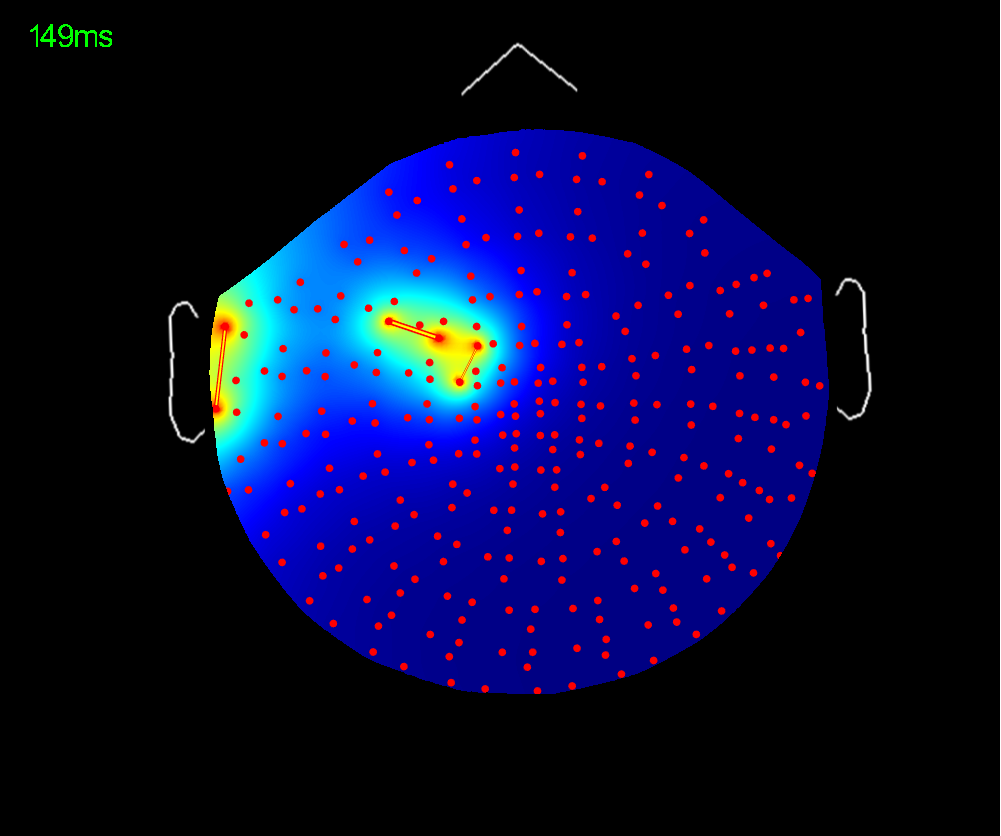} \tabularnewline
(a) & (b) & (c) & (d) \tabularnewline
\end{tabular}
\par\end{centering}
\caption{Graph Estimation results. First row, the initial graph. Second row, (a) and (b) are the results for the propose method and GL-SigRep, respectively, on signal data from 96ms - 105ms. (c) and (d) are the results for the proposed method and GL-SigRep, respectively, on signal data from 141ms - 150ms.}
\label{fig:BrainFigure}
\end{figure}

For another comparison, in Figure \ref{fig:BrainFigure}, we visualize the estimated graph matrix $\Phi_f$ by our method as well as the GL-SigRep method, by projecting the graph connectivity weights on the MEG sensor locations. The first row of Figure \ref{fig:BrainFigure} shows the initialization graph used for both method, the coherence connectivity graph obtained from the resting state data. The second row visualizes the output of the two methods for data at 100ms and 145ms time point.

Comparing the graph visualization results from GL-SigRep (Figure \ref{fig:BrainFigure}, b and d), one can see that using the data from 96-105ms, GL-SigRep indicates connectivities at the left temporal and middle and inferior frontal gyri. The estimated graph by GL-SigRep does not significantly change using the data from 141-150ms either.  None of these regions high-lighted by GL-SigRep has apparent link with early visual processing described in neuroscientific literature (note that GL-SigRep assumes Gaussian noise and would not be appropriate for MEG data).
At 100ms after projection of a visual stimuli like a face image, neuroscientific literature seem to report the visual information is still being processed at early visual cortex at occipital and occipitotemporal regions (e.g. see \cite{Thorpe1996,Desimone2006}). Unlike GL-SigRep results, at 96-105ms, the graph connectivity estimation by our method (Figure \ref{fig:BrainFigure}, a) tends to span more on the occipital and left occipitotemporal regions, and therefore seem to have reached a more successful estimation of the true underlying connectivity at this time-point.

The connectivity graph estimated by our method during 141ms to 150ms (Figure \ref{fig:BrainFigure}, c) converges on connections on the right occipitotemporal region. This graph connectivity is comparable to the neuroscientific findings on face perception, and specifically the N170 marker. In several studies such as \cite{Perrett1987,Liu2002}, the fusiform gyrus (at the occipitotemporal region) are suggested for processing face perception during about 145ms after presentation of a face image stimuli, also known as N170 marker (named after its first discovery at 170ms in EEG studies). In this work, our technique reveals almost the same regions as {\em important} graph connections for face perception. The compatibility of our estimated graph connectivity with neuroscientific literature further supports our proposed method over others.

\begin{table}\small 
	\begin{tabular}{|*{1}{p{70pt}|}*{4}{c|}}
		\hline
		Time slot&\multicolumn{2}{c|}{96ms-105ms}&\multicolumn{2}{c|}{141ms-150ms}\\
		\hline
		Methods&SVM&rank&SVM&rank \\
		\hline
		Proposed Method& \textbf{66.65\%} & 21 & \textbf{81.91\%} & 25\\
		\hline
		GL-SigRep&64.97\%&21 &78.93\%&25\\
		\hline
		RPCAG&64.21\%&23 &81.15\%&29\\
		\hline
		RPCA&62.89\%&32 &80.15\%& 33\\		
		\hline
		PCA&64.22\%&32 &78.81\%& 33\\
		\hline
	\end{tabular}
	\centering
	\vspace{+0.1cm}
	\normalsize
	\caption{Classification performance (accuracies) for brain imaging data in two different time slots.}
	\label{tab:BrainResult}
\end{table}

\section{Conclusion}
\label{sec:conclusion}

We propose an algorithm in learning the low rank component and graph simultaneously. It is suitable for cases where the perturbations on the low rank components are grossly but sparse. We showed that the proposed method on both synthetic data and brain imaging data is competitive. Our method achieves good performance on both low rank approximation and graph estimation. In addition, when applying to the brain imaging data, our method could recover a connectivity graph that is more compatible to the neuroscientific literature, indicating its better estimation of the true underlying graph. Future work applies the proposed algorithm for other high-dimensional data~\cite{fang:tip:2014}.




\vfill\pagebreak



\bibliographystyle{IEEEbib}
\footnotesize{
\bibliography{refs}

\begin{thebibliography}{10}

\bibitem{shuman2013emerging}
David Shuman, Sunil~K Narang, Pascal Frossard, Antonio Ortega, Pierre
  Vandergheynst, et~al.,
\newblock ``The emerging field of signal processing on graphs: Extending
  high-dimensional data analysis to networks and other irregular domains,''
\newblock {\em Signal Processing Magazine, IEEE}, vol. 30, no. 3, pp. 83--98,
  2013.

\bibitem{frossard2015}
Xiaowen Dong, Dorina Thanou, Pascal Frossard, and Pierre Vandergheynst,
\newblock ``Laplacian matrix learning for smooth graph signal representation,''
\newblock in {\em Proc. {ICASSP}}, 2015.

\bibitem{ortega:2016}
Eduardo Pavez and Antonio Ortega,
\newblock ``Generalized laplacian precision matrix estimation for graph signal
  processing,''
\newblock in {\em Proc. {ICASSP}}, 2016.

\bibitem{Kalofolias:2016}
Vassilis Kalofolias,
\newblock ``How to learn a graph from smooth signals,''
\newblock in {\em AISTATS 2016}, 2016.

\bibitem{hossein2016}
Kleovoulos Tsourides, Shahriar Shariat, Hossein Nejati, Tapan~K Gandhi, Annie
  Cardinaux, Christopher~T Simons, Ngai-Man Cheung, Vladimir Pavlovic, and
  Pawan Sinha,
\newblock ``Neural correlates of the food/non-food visual distinction,''
\newblock {\em Biological Psychology}, 2016.

\bibitem{Nejati2015}
H.~Nejati, K.~Tsourides, V.~Pomponiu, E.C. Ehrenberg, Ngai-Man Cheung, and
  P.~Sinha,
\newblock ``Towards perception awareness: Perceptual event detection for brain
  computer interfaces,''
\newblock in {\em EMBC2015}, Aug 2015, pp. 1480--1483.

\bibitem{bullmore2009complex}
Ed~Bullmore and Olaf Sporns,
\newblock ``Complex brain networks: graph theoretical analysis of structural
  and functional systems,''
\newblock {\em Nature Reviews Neuroscience}, vol. 10, no. 3, pp. 186--198,
  2009.

\bibitem{hyde2012cross}
James~S Hyde and Andrzej Jesmanowicz,
\newblock ``Cross-correlation: an fmri signal-processing strategy,''
\newblock {\em NeuroImage}, vol. 62, no. 2, pp. 848--851, 2012.

\bibitem{brovelli2004beta}
Andrea Brovelli, Mingzhou Ding, Anders Ledberg, Yonghong Chen, Richard
  Nakamura, and Steven~L Bressler,
\newblock ``Beta oscillations in a large-scale sensorimotor cortical network:
  directional influences revealed by granger causality,''
\newblock {\em Proceedings of the National Academy of Sciences of the United
  States of America}, vol. 101, no. 26, pp. 9849--9854, 2004.

\bibitem{boyd2011distributed}
Stephen Boyd, Neal Parikh, Eric Chu, Borja Peleato, and Jonathan Eckstein,
\newblock ``Distributed optimization and statistical learning via the
  alternating direction method of multipliers,''
\newblock {\em Foundations and Trends in Machine Learning}, vol. 3, no. 1, pp.
  1--122, 2011.

\bibitem{lu:2013}
Chenhui Hu, Lin Cheng, Jorge Sepulcre, Georges~El Fakhri, Yue~M. Lu, and
  Quanzheng Li,
\newblock ``A graph theoretical regression model for brain connectivity
  learning of alzheimer’s disease,''
\newblock in {\em ISBI2013}, San Francisco, CA, 7-11 Apr. 2013.

\bibitem{Jian13b}
Bo~Jiang, Chris Ding, Bin Luo, and Jin Tang,
\newblock ``{Graph-Laplacian PCA}: Closed-form solution and robustness,''
\newblock in {\em CVPR}, 2013, pp. 3490--34968.

\bibitem{Shah15a}
Nauman Shahid, Vassilis Kalofolias, Xavier Bresson, Michael Bronstein, and
  Pierre Vandergheynst,
\newblock ``Robust principal component analysis on graphs,''
\newblock in {\em Proceedings of International Conference on Computer Vision},
  Santiago, Chile, 2015, pp. 2812--2820.

\bibitem{Shah16b}
Nauman Shahid, Nathanael Perraudin, Vassilis Kalofolias, Gilles Puy, and Pierre
  Vandergheynst,
\newblock ``Fast robust {PCA} on graphs,''
\newblock {\em arXiv:1507.08173v2 [cs.CV] 25 Jan 2016}, pp. 1--17, 2016.

\bibitem{huang2016graph}
Weiyu Huang, Leah Goldsberry, Nicholas~F Wymbs, Scott~T Grafton, Danielle~S
  Bassett, and Alejandro Ribeiro,
\newblock ``Graph frequency analysis of brain signals,''
\newblock {\em arXiv preprint arXiv:1512.00037v2}, 2016.

\bibitem{rui2016dimensionality}
Liu Rui, Hossein Nejati, and Ngai-Man Cheung,
\newblock ``Dimensionality reduction of brain imaging data using graph signal
  processing,''
\newblock in {\em ICIP}. IEEE, 2016, pp. 1329--1333.

\bibitem{behjat2015anatomically}
Hamid Behjat, Nora Leonardi, Leif S{\"o}rnmo, and Dimitri Van De~Ville,
\newblock ``Anatomically-adapted graph wavelets for improved group-level fmri
  activation mapping,''
\newblock {\em NeuroImage}, vol. 123, pp. 185--199, 2015.

\bibitem{hu2015multiresolution}
Wei Hu, Gene Cheung, Antonio Ortega, and Oscar~C Au,
\newblock ``Multiresolution graph fourier transform for compression of
  piecewise smooth images,''
\newblock {\em Image Processing, IEEE Transactions on}, vol. 24, no. 1, pp.
  419--433, 2015.

\bibitem{egilmez2014spectral}
Hilmi~E Egilmez and Antonio Ortega,
\newblock ``Spectral anomaly detection using graph-based filtering for wireless
  sensor networks,''
\newblock in {\em ICASSP}. IEEE, 2014, pp. 1085--1089.

\bibitem{dong2013inference}
Xiaowen Dong, Antonio Ortega, Pascal Frossard, and Pierre Vandergheynst,
\newblock ``Inference of mobility patterns via spectral graph wavelets,''
\newblock in {\em ICASSP}. IEEE, 2013, pp. 3118--3122.

\bibitem{kang2014complex}
Jieqi Kang, Shan Lu, Weibo Gong, and Patrick~A Kelly,
\newblock ``A complex network based feature extraction for image retrieval,''
\newblock in {\em ICIP}. IEEE, 2014, pp. 2051--2055.

\bibitem{Belk03}
M.~Belkin and P.~Niyogi,
\newblock ``Laplacian eigenmaps for dimensionality reduction and data
  representation,''
\newblock {\em Neural Computation}, vol. 15, no. 6, pp. 1373--1396, 2003.

\bibitem{He:2003}
Xiaofei He and Partha Niyogi,
\newblock ``Locality preserving projection,''
\newblock in {\em Proceedings of NIPS}, 2003.

\bibitem{yan:2007}
Shuicheng Yan, Dong Xu, Benyu Zhang, and Stephen Lin,
\newblock ``Graph embedding and extensions: a general framework for
  dimensionality reduction,''
\newblock {\em IEEE Transactions on Pattern Analysis and Machine Intelligence},
  2007.

\bibitem{Cand11}
Emmanuel~J. Candès, Xiaodong Li, Yi~Ma, and John Wright,
\newblock ``Robust principal component analysis?,''
\newblock {\em Journal of the ACM}, vol. 58, no. 3, pp. 11:1--11:37, May 2011.

\bibitem{dong2014learning}
Xiaowen Dong, Dorina Thanou, Pascal Frossard, and Pierre Vandergheynst,
\newblock ``Learning laplacian matrix in smooth graph signal representations,''
\newblock {\em arXiv preprint arXiv:1406.7842v3}, 2014.

\bibitem{schmidt2014whole}
Benjamin~T Schmidt, Avniel~S Ghuman, and Theodore~J Huppert,
\newblock ``Whole brain functional connectivity using phase locking measures of
  resting state magnetoencephalography,''
\newblock {\em Front. Neurosci}, vol. 8, no. 141, pp. 10--3389, 2014.

\bibitem{Perrett1987}
David~I. Perrett, Amanda~J. Mistlin, and Andrew~J. Chitty,
\newblock ``Visual neurones responsive to faces,''
\newblock {\em Trends in Neurosciences}, vol. 10, no. 9, pp. 358 -- 364, 1987.

\bibitem{Thorpe1996}
S.~Thorpe, D.~Fize, and C.~Marlot,
\newblock ``{Speed of processing in the human visual system},''
\newblock {\em Nature}, vol. 381, no. 6582, pp. 520--2, 1996.

\bibitem{Liu2002}
Jia Liu, Alison Harris, and Nancy Kanwisher,
\newblock ``{Stages of processing in face perception: an MEG study.},''
\newblock {\em Nature neuroscience}, vol. 5, no. 9, pp. 910--916, Sept. 2002.

\bibitem{Desimone2006}
R.~Desimone,
\newblock ``Face-selective cells in the temporal cortex of monkeys,''
\newblock {\em Journal of Cognitive Neuroscience}, vol. 3, no. 1, pp. 1--8,
  Jan. 2006.

\bibitem{fang:tip:2014}
Lu~Fang, Ngai-Man Cheung, D~Tian, A~Vetro, H~Sun, and O~Au,
\newblock ``An analytical model for synthesis distortion estimation in 3d
  video,''
\newblock {\em IEEE Transactions on Image Processing}, vol. 23, no. 1, pp.
  185--199, 2014.

\end{thebibliography}
}

\end{document}